\pdfoutput=1
\documentclass[11pt]{article}
\usepackage[]{EACL2023}

\usepackage{times}
\usepackage{latexsym}
\usepackage[T1]{fontenc}
\usepackage[utf8]{inputenc}
\usepackage{microtype}
\usepackage{inconsolata}
\usepackage{amsmath} 
\usepackage{amssymb}
\usepackage{graphicx}
\usepackage{pdfpages}
\usepackage{booktabs} 
\usepackage{multirow} 
\usepackage{tabularx}

\title{Next Visit Diagnosis Prediction via Medical Code-Centric \\ Multimodal Contrastive EHR Modelling with Hierarchical Regularisation}

\author{
    Heejoon Koo \\
    University College London \\
    heejoon.koo.17@alumni.ucl.ac.uk
}

\begin{document}
\maketitle

\begin{abstract}

Predicting next visit diagnosis using Electronic Health Records (EHR) is an essential task in healthcare, critical for devising proactive future plans for both healthcare providers and patients. Nonetheless, many preceding studies have not sufficiently addressed the heterogeneous and hierarchical characteristics inherent in EHR data, inevitably leading to sub-optimal performance. To this end, we propose NECHO, a novel medical code-centric multimodal contrastive EHR learning framework with hierarchical regularisation. First, we integrate multifaceted information encompassing medical codes, demographics, and clinical notes using a tailored network design and a pair of bimodal contrastive losses, all of which pivot around a medical codes representation. We also regularise modality-specific encoders using a parental level information in medical ontology to learn hierarchical structure of EHR data. A series of experiments on MIMIC-III data demonstrates effectiveness of our approach.

\end{abstract}

\section{Introduction}

Predicting a patient's future diagnosis has been a longstanding objective in both academic and industrial healthcare sectors. Its significance is highlighted for healthcare providers with refining decision-making processes and resource allocation, and also for patients with effective future plans. By leveraging the extensive accumulation of EHR data, data-driven deep learning methodologies have achieved considerable advancements in the healthcare practices, particularly in next admissions diagnosis prediction \citep{choi2016doctor, ma2018kame, qiao2019mnn, zhang2020combining}.

However, most of previous studies have shown limited consideration into multifaceted and hierarchical properties inherent in EHR data. First, it is heterogeneous, encompassing a range of modalities including demographics (e.g. age), medical images (e.g., Computed Tomography), text (e.g. clinical notes), time series (e.g. laboratory tests), and medical codes (e.g. ICD-9). Each modality offers diverse and unique perspectives of a single observation and holds substantial potential to improve representative power if it is integrated seamlessly with other modalities. Nevertheless, the majority of previous works have solely focused on medical codes or shown limited exploration into effective multimodal fusion strategies \citep{choi2017gram, zhang2020combining, yang2021leverage}. 

\begin{figure}[t!]
    \centering
    \includegraphics[width=0.48\textwidth]{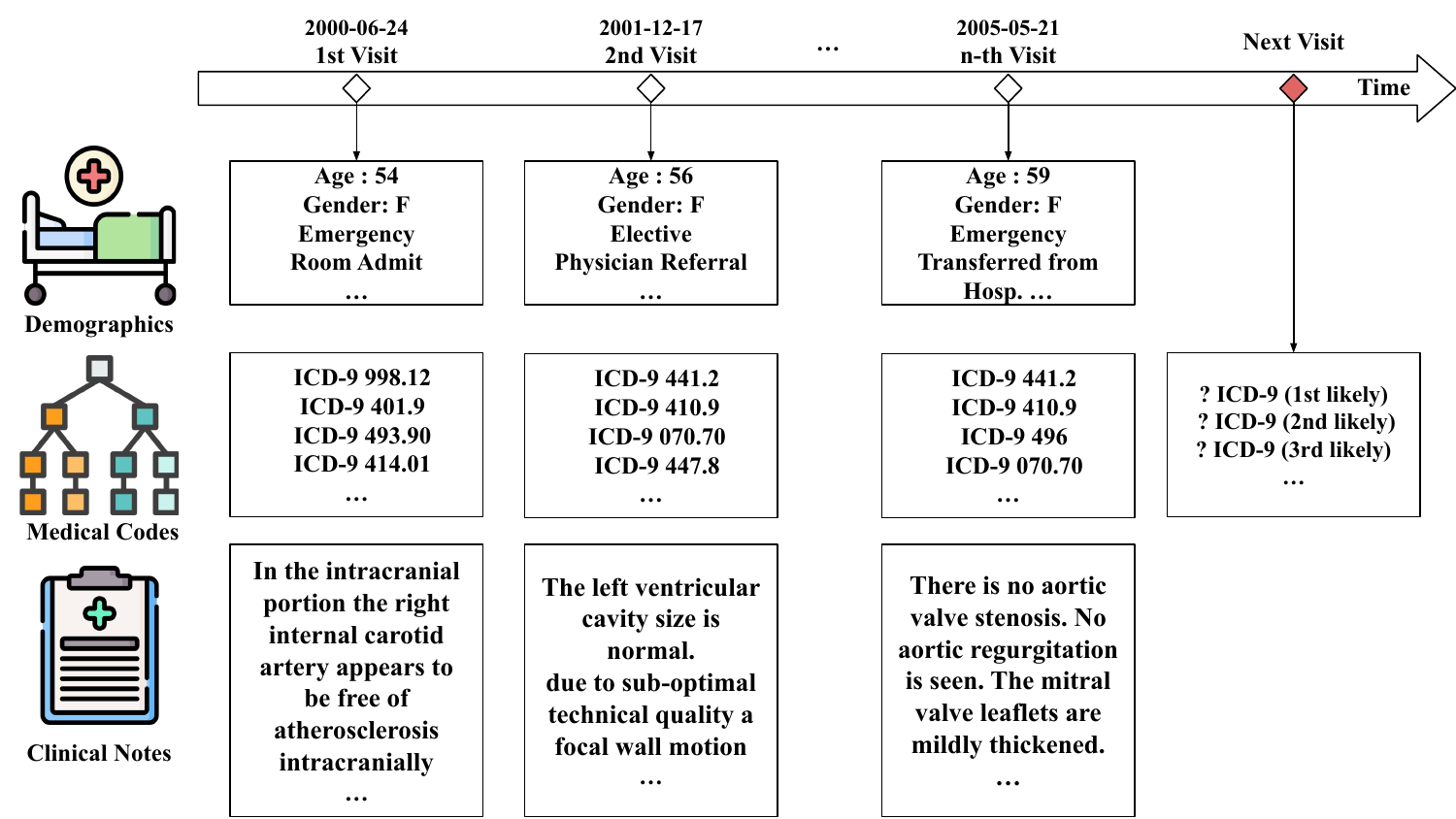}
    \caption{A Segment of Longitudinal EHR Data. It includes demographics, medical codes and clinical notes.}
    \label{fig:ehr}
\end{figure}

Second, EHR data employs International Classification of Diseases (ICD) codes \citep{slee1978international}, an organised hierarchical medical concept ontology. It is used by domain experts to systematically categorise patient diagnoses into relevant medical concepts. For instance, in its ninth version (ICD-9), circulatory system diseases (ICD-9 code 390-459) are further categorised into 9 subcategories, each denoting specific conditions, such as hypertensive disease (ICD-9 code 401-405). Each is further divided into 10 subcategories (e.g. ICD-9 code 401.0 to 401.9). This shows a highly structured and hierarchical dependency amongst them. Despite the critical importance of these attributes, they have been largely overlooked in earlier studies. 

To address the aforementioned characteristics of EHR data, we present a novel framework for \textbf{Ne}xt Visit Diagnosis Prediction via Medical Code-Centric Multimodal \textbf{C}ontrastive EHR Modelling with \textbf{H}ierarchical Regularisati\textbf{o}n (\textbf{NECHO}). To the best of our knowledge, this framework is the first work designed in a medical code-centric fashion for diagnosis prediction. It tightly and seamlessly entangles three distinct modalities of medical codes, demographics, and clinical notes through a meticulously designed multimodal fusion network and two bimodal contrastive losses. Its goal is to boost representational power by positioning demographics and clinical notes as supplementary modalities. Furthermore, we harness an auxiliary loss to regularise each modality-specialised encoder based on the ancestral level of medical codes, thereby successfully injecting more general information from the ICD-9 medical ontology. Therefore, the main contributions of our work are threefold as follows: 

\begin{itemize}
    \item We effectively integrate three distinct modalities by developing a novel fusion network and a pair of bimodal contrastive losses, centralised around medical codes representation. 
    \item We also propose to use auxiliary losses for each modality-specific model to regularise them using the parental level of medical codes to learn more general information, leveraging hierarchical nature of ICD-9 codes. 
    \item Our proposed NECHO framework achieves superior performance over previous works on MIMIC-III \citep{johnson2016mimic}, a publicly available large-scale real-world healthcare data. 
\end{itemize}

\section{Related Works}
\subsection{Next Visit Diagnosis Prediction}

AI research community has delved into future diagnosis predictions, employing various data modalities such as graph, text, or more than two. DoctorAI \citep{choi2016doctor} is the first work that predicts diagnoses utilising a simple recurrent neural networks (RNN). It is further refined to RETAIN \citep{choi2016retain} and Dipole \citep{ma2017dipole}, which incorporate attention mechanisms. 

Meanwhile, graph neural networks (GNN) have been influential, with models like GRAM \citep{choi2017gram} and KAME \citep{ma2018kame} constructing disease graphs from medical ontology, and others like MMORE \citep{song2019medical} and HAP \citep{zhang2020hierarchical} focusing on learning both ontology and diagnosis co-occurrence and leveraging hierarchical attention, respectively. MIPO \citep{peng2021mipo} predicts parental level medical codes based on the medical ontology additionally. 

Biomedical domain specific pre-trained word2vec \citep{zhang2019biowordvec} and language models have been introduced \citep{alsentzer2019publicly} for clinical text understanding. The importance of them is particularly underscored in multimodal EHR learning \citep{husmann2022importance}, often supplementing diverse prediction tasks. MNN \citep{qiao2019mnn} and CGL \citep{lu2021collaborative} fuse medical codes and clinical notes. MAIN \citep{an2021main} further integrates demographics to learn more comprehensive information of patients. \citep{yang2021leverage} explore multiple fusion strategies for clinical event prediction.

\subsection{Multimodal Learning}

Beyond EHR, multimodality learning has been explored to various domains, particularly in multimodal sentiment analysis (MSA) \citep{gandhi2022multimodal}. We introduce a few works that have somewhat influenced our work.

First, Tensor Fusion Network (TFN) \citep{zadeh2017tensor, liu2018efficient} and Multimodal Adaptation Gate (MAG) \citep{rahman2020integrating} perform an outer product and attentional gate on representations from varying modalities, respectively. \citep{tsai2019multimodal} use cross-modal and self-attention transformers \citep{vaswani2017attention}. \cite{yu2021learning} introduce Unimodal Label Generation Module (ULGM) to boost modality-wise representations. However, the above literature do not consider the modality imbalance, such as the superiority of text-based models. Based on such findings, text-centred multimodal fusion strategies have been developed \citep{qiu2022intermulti, huang2023tefna}. 

\subsection{Contrastive Learning}

Contrastive Learning has emerged as a predominant paradigm, showing its superior performance in many research areas recently. Originally, it aims to learn features from different views of a single sample and discriminate samples from different classes \citep{oord2018representation, chen2020simple}. Next, it is extended to multimodality. CLIP \citep{radford2021learning} is a seminal work on multimodal contrastive learning, employing InfoNCE loss \citep{oord2018representation} to learn transferable features between images and texts. \citep{zhang2022contrastive} apply this strategy to medical domain, whilst \citep{mai2022hybrid} exploit trimodal contrastive learning in MSA.

\begin{figure*}[t!]
    \centering
    \includegraphics[width=1.0\textwidth, angle=0]{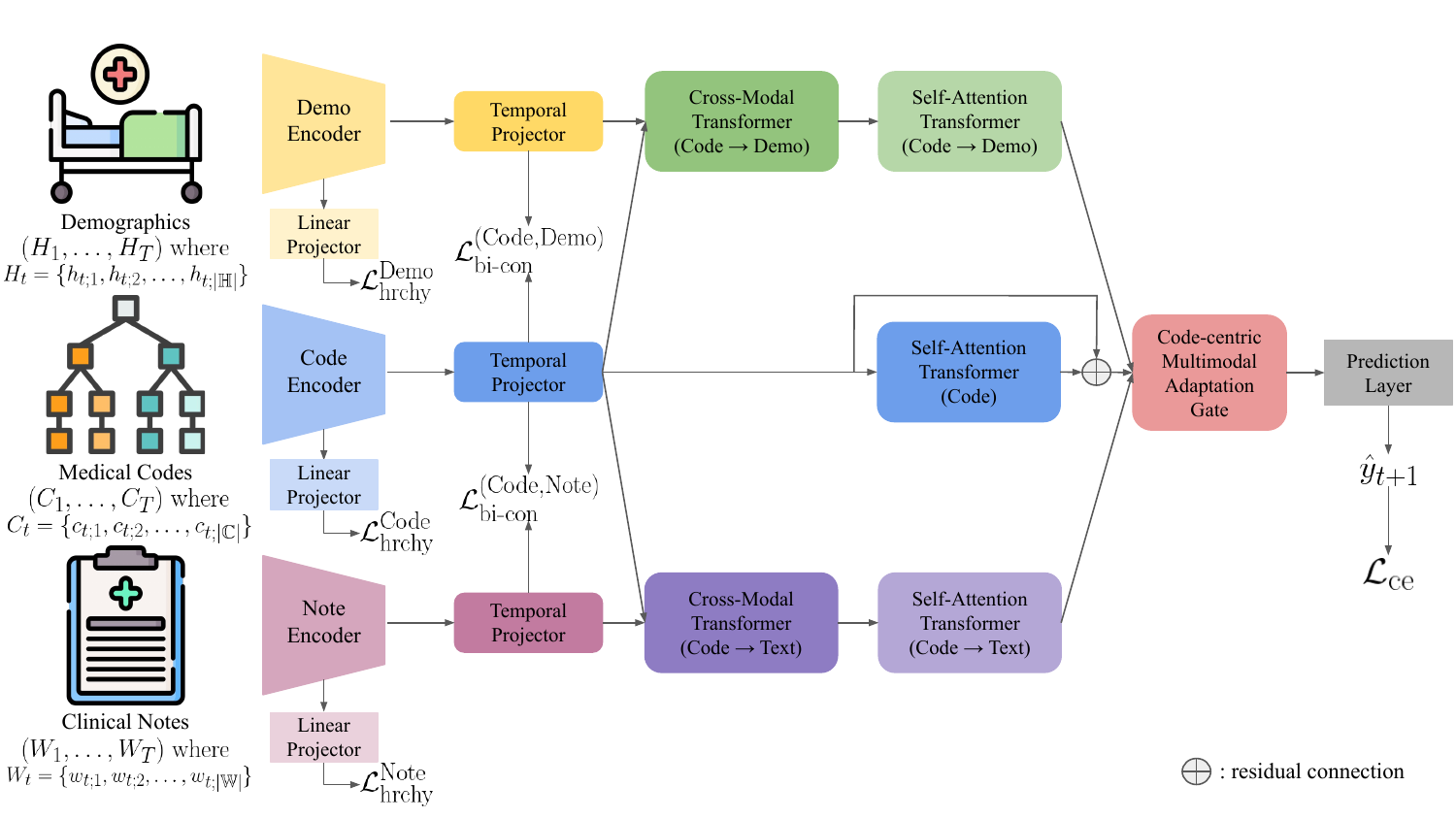}
    \caption{The Overall Framework of Our Proposed NECHO.}
    \label{fig:main_framework}
\end{figure*}

\section{Methodology}

In this section, we firstly introduce notations and problem formulation on next visit diagnosis prediction. Thereafter, we describe an overview and details of our proposed framework, NECHO. 

\subsection{Problem Formulation}

\textbf{Multimodal EHR Data} A clinical record can be represented as a time-ordered sequence of visits $ V_1, \ldots, V_T $, where $ T $ is the total number of visits of any patient $ \mathcal{P} $. Each visit $ V_t $ is denoted as $ {(C_t, A_t, H_t, W_t)} $, where $ C_t $ is a set of diagnosis codes, $ A_t $ is a set of diagnosis codes at their ancestral level, $ H_t $ is demographics, $ W_t $ is a clinical note at $ t $-th admission, respectively. 

We denote a set of medical codes from EHR data as $ c_1, c_2, \ldots, c_{\mathbb{C}} \in \mathbb{C} $, where $ \vert \mathbb{C} \vert $ is the number of unique medical codes at a level in ICD-9 code hierarchy $\mathcal{G}$. Similarly, a set of medical codes at their direct ancestral level is denoted as $ a_1, a_2, \ldots, a_{\mathbb{A}} \in \mathbb{A} $. The total number of unique medical codes in parental level is $ \vert \mathbb{A} \vert $. Note that, $ \vert \mathbb{A} \vert $  $ \ll $ $ \vert \mathbb{C} \vert $.

Diagnosis code at $t$-th visit is represented by $ C_t = \{c_{t;1}, c_{t;2}, \ldots, c_{t;\vert \mathbb{C} \vert}\} $, where $ \vert \mathbb{C} \vert $ represents the number of diagnosis codes. Its ancestral level code is denoted by $ A_t = \{a_{t;1}, a_{t;2}, \ldots, a_{t;\vert \mathbb{A} \vert}\} $ with of the number of parental level diagnosis codes $ \vert \mathbb{A} \vert $.  Demographics is represented as $ H_t = \{h_{t;1}, h_{t;2}, \ldots, h_{t;\vert \mathbb{H} \vert}\} $, where $ \vert \mathbb{H} \vert $ is the total number of demographics features. Clinical note is represented as $ W_t = \{w_{t;1}, w_{t;2}, \ldots, w_{t;\vert \mathbb{W} \vert}\} $, where $ \vert \mathbb{W} \vert $ is the maximum number of words to process. 

\textbf{Next Visit Diagnosis Prediction Task} Based on the above notations, next visit diagnosis prediction is defined as follows. Given the patient's multifaceted clinical records for the previous $ T $ visits, the objective is to predict a $ (T+1) $-th visit's diagnosis codes, denoted as $ \hat{y}_{T+1} $. 

\subsection{Medical Code Information Centred Multimodal Fusion}

One of the major challenges in the realm of AI healthcare is how to integrate the multifaceted data effectively. This has catalysed a surge of research on multimodal EHR learning \cite{zhang2020combining, yang2021leverage}. Nonetheless, a notable limitation in prior studies is the oversight of modality imbalance and the adoption of a modality-symmetric strategy, resulting in an unsatisfactory performance. We empirically observe that the medical code representations show the best performance. Also, previous works on MSA prioritise text representations at the core \citep{qiu2022intermulti, huang2023tefna} due to their superiority. Based on these findings, we introduce a novel medical code-centric multimodal fusion training scheme, which encompasses a tailored multimodal fusion network and a couple of bimodal contrastive losses. 

\subsubsection{Modality-Specific Feature Extraction}

Before introducing our novel fusion strategies, we first explain modality-specific encoders that extract features from each modality. We design them as simple as possible to highlight the efficacy of our proposed fusion strategies. In other words, our framework is modular, with the potential for performance enhancement if the encoders are switched to more representative ones.

We employ a simple embedding layer for both medical codes and demographics, and a combination of BioWord2Vec \cite{zhang2019biowordvec} and 1D CNN \cite{kim2014convolutional} to process clinical notes. Subsequently, the feature vector is passed to a fully connected layer ($\text{Linear}$) connected with ReLU activation function \citep{nair2010rectified}.
\begin{equation}
\begin{aligned}
    M_t &= \text{Encoder}_{m}(m_t), \\
    \bar{M}_t &= \text{ReLU}(\text{Linear}(M_t))
\end{aligned}
\end{equation}

where $ m_t $ is a data of modality $ m \in (C, H, W) $ at $t$-th visit and $ \text{Encoder}_{m} $ is a modality-specialised encoder, passing the feature vector $ M_t $ to MLP. Finally, a modality-specific feature $ \bar{M}_t $ is yielded. Appendix \ref{sec:method-modalityspecific} provides a detailed information on how each modality-specific encoder operates. 

\subsubsection{Multimodal Fusion Network}

\textbf{Cross-Modal Transformer} After acquiring representations from all modalities, we entangle them using two cross-modal transformers (CMTs), introduced by MulT \cite{tsai2019multimodal}. It has verified its effectiveness in integrating meaningful information across different modalities. Initially, we put the each distinct representation to a temporal non-linear projector, 1D CNN:
\begin{equation}
    \hat{H}_{t}^{m} = \text{Conv1D}(\bar{M}_t)
\end{equation}

where $ \bar{M}_t $ is a representation from any modality $ m $ and $ \hat{H}_{t}^{m} $ is a resultant representation. \text{Conv1D} is equivalent to 1D CNN. Next, we introduce cross-modal attention, which facilitates the information transfer from the source modality to the target modality, e.g. medical codes $ \to $ clinical notes. 

Let two modalities as $ m_1 $ and $ m_2 $. Then, using trainable weights $ W^{(\cdot)} $ with a dimension of $ d_k $, we define the query, key and values as $ Q^{m_1} = H^{m_1} W^{Q^{m_1}} $, $ K^{m_2} = H^{m_2} W^{K^{m_2}} $, and $ V^{m_2} = H^{m_2} W^{V^{m_2}} $, respectively. The cross-modal attention, denoted as CA, from $ m_1 $ to $ m_2 $ is then:
\begin{equation}
\begin{aligned}
   Z^{m_1 \to m_2} &= \text{CA}^{m_1 \to m_2} (\hat{H}^{m_1}, \hat{H}^{m_2}) \\
   &= \text{Softmax}(\frac{Q^{m_1}(K^{m_2})^T}{\sqrt{d_k}})V^{m_2}.
\end{aligned}
\end{equation}

We omit $ t $ for brevity. CMT is an extension of the CA. It is composed of a multi-head cross-modal attention block ($ \text{MHA} $) and a Layer Normalisation layer ($ \text{LM} $) \cite{ba2016layer}. It is computed feed-forwardly for $ i = 1, \ldots, D $ layers as follows:
\begin{equation}
    Z^{m_1 \to m_2}_{(0)} = H^{m_2}_{(0)},
\end{equation}
\begin{equation}
\begin{aligned}
    \hat{Z}^{m_1 \to m_2}_{(i)} = \text{MHA}^{m_1 \to m_2}_{(i)} (\text{LM}(Z^{m_1 \to m_2}_{(i-1)}) \\ 
    \text{LM}(H^T_{(0)})) + \text{LM}(Z^{m_1 \to m_2}_{(i-1)}),
\end{aligned}
\end{equation}
\begin{equation}
\begin{aligned}
    Z^{m_1 \to m_2}_{(i)} = f_{\theta^{m_1 \to m_2}_{(i)}} (\text{LM}(\hat{Z}^{m_1 \to m_2}_{(i)})) + \\
    \text{LM}(\hat{Z}^{m_1 \to m_2}_{(i)}).
\end{aligned}
\end{equation}

During the process at MHA, the representations from the source modality are correlated with the target modality, enhancing the representational power across different modalities. As presented in Fig. \ref{fig:main_framework}, the fusion is performed in a medical code-centric fashion, thus we set $ m_1 $ as medical code $ C $ and $ m_2 $ as either demographics $ H $ or clinical notes $ W $. Thus, we acquire two representations of $ Z^{C \to H}_t $ and $ Z^{C \to W}_t $ from the two CMTs.

\textbf{Self-Attention Transformer} To extract sequential feature representations effectively and boost dependencies from the above two cross-modal and medical code representations, a self-attention transformer ($ \text{SA} $) is employed. It processes across the patient's single visits:
\begin{equation}
\begin{aligned}
   \hat{y}^C_t &= \text{SA}^C(\hat{H}^C_t), \\
   \hat{y}^{C \to H}_t &= \text{SA}^{C \to H}(Z^{C \to H}_t), \\
   \hat{y}^{C \to W}_t &= \text{SA}^{C \to W}(Z^{C \to W}_t). \\
\end{aligned}
\end{equation}

Additionally, we perform a residual connection \cite{he2016deep} between the code representation before and after $ \text{SA}^C $ to enhance the influence of the medical code modality representation. 
\begin{equation}
    \hat{y}_t^C = \hat{y}_t^C + \hat{H}_{t}^C.
\end{equation}

\textbf{Multimodal Adaptation Gate} Rather than performing a simple concatenation of the three distinct representations, we modify and adopt previous multimodal adaptation gate (MAG) \cite{rahman2020integrating, yang2021leverage} in the medical code-centric manner. First, we calculate the trimodal gating value $ g \in \mathbb{R} $ and the displacement vector $ \text{H} $ by concatenating meaningful representations in the previous stage as:
\begin{equation}
    g = \text{Linear}(\text{concat}(\hat{y}_t^C ; \hat{y}_t^{C \to H} ; \hat{y}_t^{C \to W})), 
\end{equation}
\begin{equation}
    \text{H} = \text{Linear}(g(\text{concat}(\hat{y}_t^{C \to H} ; \hat{y}_t^{C \to W}))).
\end{equation}

This modification maximises the influence of medical code representation during the multimodal fusion process. Then, a weighted summation is performed between the medical code representation $ \hat{y}_t^C $ and the displacement vector $ \text{H} $ to derive the multimodal representation $ \text{M} $:
\begin{equation}
\begin{aligned}
    \text{M} &= \hat{y}_t^C + \alpha \text{H}, \\
    \text{where} \hspace{0.1cm} \alpha &= \text{min}(\frac{ \left\Vert \hat{y}_t^C \right\Vert_2}{ \left\Vert \text{H} \right\Vert_2} \beta, 1).
\end{aligned}
\end{equation}

Here, $ \alpha $ is a scaling factor, modulating the influence of the displacement vector $ \text{H} $ and $ \beta $ is a trainable parameter that is randomly initialised. Both $ \left\Vert \hat{y}_t^C \right\Vert_2 $ and $ \left\Vert \text{H} \right\Vert_2 $ are the $ L_2 $ norm of their respective entities. Finally, we apply a layer normalisation and dropout to $ \text{M} $. 

\textbf{Prediction} To predict next visit diagnosis, we feed the representation $ \text{M} $ in the previous stage into a single linear layer with a Sigmoid activation function to calculate the predicted probability $ \hat{y}_{t+1} $.
\begin{equation}
    \hat{y}_{t+1} = \text{Sigmoid}(\text{Linear}(\text{M})),
\end{equation}
\begin{equation}
\begin{aligned}
    \mathcal{L}_{\text{{ce}}} &= \frac{1}{T} \sum_{t=1}^{T} - \left( {y}_{t+1}^{\text{{T}}} \log \hat{{y}}_{t+1} + \right. \\
    &\quad \left. (1-{y}_{t+1})^{\text{{T}}} \log(1-\hat{{y}}_{t+1}) \right)
\end{aligned}
\end{equation}

where cross-entropy loss $ \mathcal{L}_{\text{{ce}}} $ is applied as the loss function. $ y_{t+1} $ is a ground truth with elements $ \vert \mathbb{C} \vert $, which takes a value of 1 if the $i$-th code exists in $ V_{t+1} $, otherwise 0.

\subsubsection{Bimodal Contrastive Losses}

Contrastive learning has been leveraged in multimodal pre-training literature \citep{radford2021learning, zhang2022contrastive} to align diverse modalities effectively. Inspired by prior works, we apply two bimodal contrastive losses to further intricately entangle the different modalities by anchoring on the medical code representations.

Again, let two distinct modalities of $m_1$ and $m_2$, where representation vectors derived from each modality be $ \hat{H}_{i}^{m_1} $ and $ \hat{H}_{i}^{m_2} $. Given a $i$-th pair of $ (\hat{H}_{i}^{m_1}, \hat{H}_{i}^{m_2}) $, our bimodal contrastive loss scheme incorporates two asymmetric losses, $ m_1 $-to-$ m_2 $ contrastive loss for the $i$-th pair and its inverse.
\begin{equation}
    l_i^{(m_1 \to m_2)}=- \log \frac{\exp(\langle \hat{H}_{i}^{m_1}, \hat{H}_{i}^{m_2} \rangle/ \tau)}{\sum_{k=1}^N \exp(\langle \hat{H}_{i}^{m_1}, \hat{H}_{k}^{m_2}\rangle / \tau)},
\end{equation}
\begin{equation}
    l_i^{(m_2 \to m_1)}=- \log \frac{\exp(\langle \hat{H}_{i}^{m_2}, \hat{H}_{i}^{m_1} \rangle/ \tau)}{\sum_{k=1}^N \exp(\langle \hat{H}_{i}^{m_2}, \hat{H}_{k}^{m_1}\rangle / \tau)}
\end{equation}

where $ \langle, \rangle $ is cosine similarity and temperature $ \tau \in \mathbb{R}^+ $ is a parameter modulating distribution's concentration and Softmax function's gradient. Subsequently, a bimodal contrastive loss is determined by a weighted combination of $ l_i^{(m_1 \to m_2)} $ and $ l_i^{(m_2 \to m_1)} $ using a weighting parameter $\alpha \in [0,1]$ and averaging over the mini-batch $ N $ as:
\begin{equation}
\begin{aligned}
    \mathcal{L}^{(m_1, m_2)}_{\text{bi-con}} = \frac{1}{N} \sum_{i=1}^{N}(\alpha l_i^{(m_1 \to m_2)} + \\ (1 - \alpha) l_i^{(m_2 \to m_1)}).
\end{aligned}
\end{equation}

We apply this to two pairs, one between medical codes and demographics, and the other between medical codes and clinical notes.
\begin{equation}
    \mathcal{L}_{\text{bi-con}} = \mathcal{L}^{(C, H)}_{\text{bi-con}} + \mathcal{L}^{(C, W)}_{\text{bi-con}}.
\end{equation}

Note that, our multimodal contrastive loss is applied inter-modally, in line with the CLIP \cite{radford2021learning}, rather than intra-modally. Moreover, we consider at the patient level rather than at the visit level. This is because patient level representations share similar patterns between their visits.

\subsection{Hierarchical Regularisation}

Medical ontologies organise diseases in a hierarchical manner. By effectively leveraging this, models are capable of acquiring knowledge at both general and specific levels of medical codes. This approach also mitigates the risk of error propagation and minimises the loss of pertinent information throughout the intricate multimodal fusion processes.

In ULGM \citep{yu2021learning}, modality-tailored encoders are also tasked with predicting ground truths. Meanwhile, MIPO \citep{peng2021mipo} introduces an auxiliary loss to learn parental level ICD-9 code prediction. Inspired by them, we introduce a regularisation strategy for each modality-specialised encoder to learn parental level of ICD-9 codes.

Specifically, the modality-specific features $ \bar{M}_t $ are passed to fully connected layers and Sigmoid activation function, yielding modality-specific parental level prediction $ \hat{{o}}^{m}_{t} $. Subsequently, we employ three cross-entropy losses, denoted as $ {L}^{m}_{\text{{hrchy}}} $, to each modality $ m $ for this auxiliary task:
\begin{equation}
    \hat{{o}}^{m}_{t+1} = \text{Sigmoid}(\text{Linear}(\bar{M}_t)),
\end{equation}
\begin{equation}
\begin{aligned}
    \mathcal{L}^{m}_{\text{{hrchy}}} &= \frac{1}{T} \sum_{t=1}^{T} - \left( {o}_{t+1}^{\text{{T}}} \log \hat{{o}}^{m}_{t+1} + \right. \\
    &\quad \left. (1-{o}_{t+1})^{\text{{T}}} \log(1-\hat{{o}}^{m}_{t+1}) \right)
\end{aligned}
\end{equation}

$ o_{t+1} $ is a ground truth with elements $ \vert \mathbb{A} \vert $, where 1 is assigned if the $i$-th code presents in $ V_{t+1} $ and 0 if absent. This is re-written to encompass three distinct modalities as:
\begin{equation}
    \mathcal{L}_{\text{hrchy}} = \mathcal{L}^{C}_{\text{{hrchy}}} + \mathcal{L}^{H}_{\text{{hrchy}}} + \mathcal{L}^{W}_{\text{{hrchy}}}.
\end{equation}

\subsection{Model Optimisation} 

The final objective function $ \mathcal{L}_{\text{total}} $ is a weighted sum of three loss terms: the cross-entropy loss $ \mathcal{L}_{\text{ce}} $ between ground truth diagnosis and prediction, the medical code-centric two bimodal contrastive losses $ \mathcal{L}_{\text{bi-con}} $, and the three modality-specific direct ancestral level hierarchical losses $ \mathcal{L}_{\text{hrchy}} $. It is formulated as:
\begin{equation}
    \mathcal{L}_{\text{total}} = \lambda_{\text{ce}} \mathcal{L}_{\text{ce}} + \lambda_{\text{bi-con}} \mathcal{L}_{\text{bi-con}} + \lambda_{\text{hrchy}} \mathcal{L}_{\text{hrchy}}
\end{equation}

where $ \lambda_{\text{ce}}, \lambda_{\text{bi-con}} $, and $ \lambda_{\text{hrchy}} $ are parameters that balance the different loss terms. The parameters of the model are updated via stochastic gradient descent (SGD) technique with respect to the calculated loss.

\begin{table*}[t]
\centering
\resizebox{0.9\textwidth}{!}{%
\begin{tabular}{cccccccc}
\toprule
\midrule
\multirow{2}{*}{\textbf{Criteria}} & \multirow{2}{*}{\textbf{Modalities}} & \multirow{2}{*}{\textbf{Models}} & \multicolumn{4}{c}{\textbf{Acc@$\textbf{k}$}}   \\
\cmidrule(lr){4-7}
      &  &          & \textbf{5}            & \textbf{10}           & \textbf{20}          & \textbf{30}          \\ 
\midrule 
\midrule
\multirow{15}{*}{EHR Modelling} & \multirow{5}{*}{Code}   & GRAM \citep{choi2017gram} & 24.16    & 36.47        & 52.48        & 62.76      \\
   &  & KAME \citep{ma2018kame}   & 25.34    & 36.93        & 54.25        & 64.54      \\
   &  & MMORE \citep{song2019medical}   & 25.97    & 38.58        & 57.05        & 68.23      \\    
   &  & MIPO \citep{peng2021mipo}   & 28.70    & \textbf{43.98}     & 60.85        & 71.07      \\    
   &  & Code Extractor (Ours)   & 28.16    & 41.83        & 57.99        & 68.31      \\ 
\cmidrule(lr){2-7}
& Demo    & Demo Extractor (Ours)  & 17.96    & 29.58       & 47.13        & 58.94      \\ 
\cmidrule(lr){2-7}
   &  \multirow{3}{*}{Note}  & BioWord2Vec $_{\text{10k}}$ \citep{zhang2019biowordvec}    & 27.31    & 41.14        & 58.53        & 69.21      \\ 
  & & BioWord2Vec $_{\text{512}}$ \citep{zhang2019biowordvec}     & 23.05    & 35.74        & 52.76        & 63.20      \\ 
   &  & Clinical BERT $_{\text{512}}$ \citep{alsentzer2019publicly}    & 24.63    & 37.21        & 54.96        & 66.37      \\\cmidrule(lr){2-7}
& Code + Note    & MNN \citep{qiao2019mnn}    & 28.16    & 41.83        & 59.75        & 69.44      \\ 
\cmidrule(lr){2-7}
& \multirow{5}{*}{Code + Demo + Note}  & MAIN \citep{an2021main}    & 27.25    & 41.07        & 57.37        & 67.69     \\ 
&  & NECHO $_{\text{w/o} \: \text{code centring} }$ (Ours)    & 28.10    & 42.13        & 59.32       & 70.01      \\
&  & NECHO $_{\text{w/o} \: \mathcal{L}_{\text{hrchy}}}$ (Ours)    & 28.71    & 43.14        & 59.83       & 70.22      \\
&  & NECHO (Ours)    & 28.66    & 43.55        & 60.77        & 71.45      \\ 
&  & NECHO $_{\text{w/} \: \text{MIPO} }$ (Ours)    & \textbf{29.05}    & 43.80        & \textbf{61.33}        & \textbf{72.08}      \\ 
\midrule
\multirow{6}{*}{Fusion Strategies} & \multirow{6}{*}{Code + Demo + Note}   & Concat & 28.38   &  42.39       &  58.63       & 68.89     \\
&    & TFN \citep{zadeh2017tensor} & 24.66   & 36.80        & 52.93        & 63.85     \\
&    & MulT \citep{tsai2019multimodal} & 28.27  &  41.87       & 58.12        & 68.50     \\
&    & MAG \citep{rahman2020integrating} & 28.26   & 42.36     & 58.40        & 69.16     \\
&    & ULGM \citep{yu2021learning} & 28.58    &  42.09      & 58.70        & 68.53     \\
&    & TeFNA \citep{huang2023tefna} & 28.12  & 41.78        &  59.11       &  69.21    \\
\bottomrule
\end{tabular}%
}
\caption{Experimental Results on MIMIC-III Data for Next Visit Diagnosis Prediction. Code, Demo, and Note are short for Medical Codes, Demographics, Clinical Notes, respectively. Best results are in boldface. 10k and 512 indicates the number of words. Unless specified otherwise, 10k words are processed for multimodal models with clinical notes.}
\label{tab-main}
\end{table*}

\section{Experiments}
\subsection{Experimental Setup}
\subsubsection{Dataset} 
We conduct experiments on a publicly available large-scale, deidentified real-world EHR data, MIMIC-III \citep{johnson2016mimic}. It is acquired from intensive care units (ICU) patients at Beth Israel Deaconess Medical Center between 2001 and 2012. It contains multifaceted data, including ICD-9 medical codes, demographics, clinical notes, and so on. We provide descriptions on data pre-processing and the corresponding statistics to Appendix \ref{sec:appendix-data}.

\subsubsection{Implementation Details}
We describe the details for implementation. First, we set 256 and 0.1 as a hidden dimension and a dropout rate across the entirety of the model (e.g. medical code and demographics feature extraction modules, Transformers including CMT and SA, and MAG), respectively. In the clinical note extraction module, filter sizes are set to [2, 3, 4], and the hidden dimension is 512. For the CMTs and SAs, we set the number of heads and encoder layers to be 4 and 3, respectively.

Also, following the previous work \citep{radford2021learning}, the temperature $ \tau $ and alpha $ \alpha $ are 0.1 and 0.25 for the contrastive loss. The coefficients of loss terms, $ \lambda_{\text{ce}}$, $\lambda_{\text{bi-con}} $, and $ \lambda_{\text{hrchy}} $ are set to 1, 1, and 0.1, respectively. Especially, the $ \lambda_{\text{hrchy}} $ is set relatively small to weakly regularise each modality-specific encoder to learn the parental levels of ICD-9 codes, without overly constraining them. We provide the experimental results on the different $ \lambda_{\text{hrchy}} $ to Appendix \ref{sec:appendix-hierarchy}.

\subsubsection{Training Details} 
We train models using Adam optimiser \citep{kingma2014adam} with a constant learning rate of 1e-4 and mini-batch size of 4, for a maximum of 50 epochs. The training is stopped if there is no gain for consecutive 5 epochs on validation data. Also, following the previous work \citep{choi2017gram}, our proposed framework is evaluated using top-$k$ accuracy, ranging $k$ from 5, 10, 20 to 30. This is consistent with how physicians consider a comprehensive set of potential diagnoses, and is suitable for multi-label classification scenarios where multiple diseases often co-occur. Details on other baselines are provided to Appendix \ref{sec:appendix-baselines}. 

Our proposed framework is implemented using PyTorch \citep{paszke2019pytorch} and accelerated via a single NVIDIA GeForce RTX 3090 GPU. 

\subsection{Experimental Results}
\subsubsection{Next Visit Diagnosis Prediction Results}

Table \ref{tab-main} provides quantitative results of the proposed NECHO in comparison to the baselines on the MIMIC-III data for the diagnosis prediction task. NECHO notably excels over all existing baselines in EHR modelling and multimodal fusion strategies. Its effectiveness is attributed to its ability to leverage unique and complementary information from other modalities, which especially improves top-30 accuracy ranging from 0.5\% to 10.7\% over modality-specific encoders that constitute NECHO. 

As shown in Table \ref{tab-main}, the multimodal fusion is imperative. It's noteworthy that whilst MAIN \cite{an2021main} employs a trimodal representation learning, its performance falls short compared to the bimodal MNN \cite{qiao2019mnn}. This discrepancy might arise from the harmful effects of improperly fusing demographic data lately. Especially, bimodal MNN shows comparable performance to trimodal fusion strategies baselines. This confirms the limitations of the tertiary symmetric multimodal fusion methodologies and raises the need for a medical code-centric approach, taking into account the modality imbalance.

To validate the efficacy of our fusion strategy, we compare NECHO that excludes the hierarchical regularisation (NECHO $_{\text{w/o} \ \mathcal{L}_{\text{hrchy}}}$) amongst multimodal EHR modelling and fusion baselines. Our method demonstrates superior performance over them, including NECHO $_{\text{w/o code centring}}$. These findings highlight the significance of designing multimodal fusion framework by centring medical codes representation that ensures a seamless aggregation of diverse data modalities. Furthermore, we also provide a comparative study on our novel code-centric MAG with others \cite{rahman2020integrating, yang2021leverage} to Appendix \ref{sec:appendix-MAG}.

Next, we delve into the significance of regularising modality-specific encoders using parental level of medical codes. We juxtapose NECHO with NECHO $_{\text{w/o} \ \mathcal{L}_{\text{hrchy}}}$ and ULGM \cite{yu2021learning}, at which modality-specific encoders learn the same level of medical ontology as the final prediction. They two show inferior performance, emphasising the importance of our novel strategy. It is discussed further in Ablation Studies (Section \ref{sec:ablation-studies}).

Furthermore, whilst NECHO does not completely surpass MIPO, replacing its simple medical code encoder with MIPO (NECHO $_{\text{w/ MIPO} }$) outperforms MIPO. It especially achieves a 1.01\% increase in top-30 accuracy, indicating that 1) our framework is modular, and 2) NECHO can predict additional accurate diseases than MIPO by leveraging complementary information from various modalities, emphasising its significance in real clinical settings. We provide a regarding case study (Section \ref{sec:case-study}).

Another noteworthy point outside the multimodal strategies is that, amongst the clinical note baselines, Clinical BERT \citep{alsentzer2019publicly} that is trained with a maximum of 512 tokens surpasses the combination model of BioWord2Vec \citep{zhang2019biowordvec} and 1D CNN \citep{kim2014convolutional} with equivalent number of tokens but is inferior to that model trained with 10k tokens. This suggests that enhancing performance is more about processing a large number of tokens than increasing model complexity in EHR learning. This also justifies our preference for BioWord2Vec over Clinical BERT within the realm of Pretrained Language Models.

\subsubsection{Ablation Studies}
\label{sec:ablation-studies}

We conduct ablation studies to discern influence of each module on the overall performance as: 1) individual modalities, 2) the multimodal fusion strategies (including Transformers, MAG, and bimodal contrastive losses), and 3) the hierarchical regularisation. The results are reported in Table \ref{tab-ablation}. 

Firstly, we assess the contribution of each modality within our proposed framework. The results demonstrate a clear superiority of the trimodal approach over its unimodal and bimodal ones. This underscores the unique representations from each modality are complementary to one another. Also, the significant performance degradation is observed upon the exclusion of medical code representation (w/o code), highlighting its pivotal role and rationalising our medical code-centred strategy. Additionally, whilst the exclusion of either notes or demographics similarly harms the performance, the note contains more meaningful information necessary than demographics, as shown in Table \ref{tab-main}.

Secondly, we evaluate the impact of our medical code-centred strategies by removing each component. The resultant performance decline highlights their importance. Intriguingly, the performance disparities between models lacking transformers (w/o Transformers), lacking MAG (w/o MAG), and the full model (NECHO) widen as the value of $ k $ increases, suggesting an amplified effect in scenarios involving a broader range of disease sampling. Conversely, the influence of contrastive losses (w/o $ \mathcal{L}_{\text{bi-con}} $) remains relatively stable across different top-$k$ accuracies, indicating that they effectively align the distinct modalities in a semantically consistent fashion. These observations show that the adaptation of the proposed modules simultaneously is essential for effective inter-modality interaction and integration, thereby yielding significant performance enhancements.

\begin{table}[t!]
\resizebox{0.48\textwidth}{!}{
\begin{tabular}{ccccc}
\toprule
\midrule
\multirow{2}{*}{\textbf{Criteria}} & \multirow{2}{*}{\textbf{Components}} & \multicolumn{2}{c}{\textbf{Acc@$\textbf{k}$}}   \\
\cmidrule(lr){3-4}
&        & \textbf{10}    & \textbf{30} \\ 
\midrule 
\midrule
\multirow{3}{*}{Modalities} & w/o Code    & 36.78    & 65.54      \\
   &  w/o Demo   & 42.56     & 70.12       \\
   &  w/o Note  & 41.94    & 69.00      \\    
\midrule
\multirow{3}{*}{Multimodal Fusion}   &  w/o Transformers     & 42.93    & 69.68         \\ 
& w/o MAG   & 42.77    & 69.48          \\ 
& w/o $ \mathcal{L}_{\text{bi-con}} $    &  42.69   & 70.84         \\
\midrule
Hierarchical Regularisation & w/o $ \mathcal{L}_{\text{hrchy}} $   & 43.14    & 70.22     \\ 
\midrule
NECHO & Full & \textbf{43.55}    & \textbf{71.45}     \\ 
\bottomrule
\end{tabular}%
}
\caption{Ablation Studies on MIMIC-III Data.}
\label{tab-ablation}
\end{table}

Finally, the effectiveness of our novel parental level hierarchical regularisation is investigated. Its omission (w/o $ \mathcal{L}_{\text{hrchy}} $) affects adversely model's accuracy across various top-$k$ accuracies. This suggests that enforcing the encoders for three distinct modalities, guided by the parental levels of medical codes using an ICD-9 hierarchy, is essential for enhancing performance as it injects the general information and thus prevents the possible transmission of erroneous information when combining representations from distinct data modalities, thereby encouraging effective and accurate training.

\begin{table*}[ht]
\centering
\resizebox{0.80\textwidth}{!}{
\begin{tabularx}{\textwidth}{cc>{\centering\arraybackslash}X}
\toprule
\midrule
\textbf{Visit} & \textbf{Modalities / Models} & \textbf{Contents} \\
\midrule
\midrule
\multirow{9}{*}{Preceding} & \multirow{2}{*}{Demo} & Age: 67, Gender: Male, Admission Type: Emergency, Admission Location: Transfer from hospital ...\\
\cmidrule(lr){2-3}
& Codes & D96, D109, D97, D131, D101, D49, D110, D53, D138, D257 \\
\cmidrule(lr){2-3}
& \multirow{6}{*}{Notes} &  ... he was taken to the Operating Room where mitral valve replacement was performed ... Discharge Diagnosis: mitral valve mass ... He experienced some visual hallucinations ... IMPRESSION: 1.  Enlarging bilateral pleural effusions. 2.  Enlarging cardiac silhouette suspicious for a pericardial effusion, echocardiographic confirmation is suggested. \\
\midrule
\multirow{4}{*}{Subsequent} & Codes & \textbf{D238}, \textbf{D53}, D130, \textbf{D106}, \textbf{D101}, \textbf{D49}, D2, D3, \textbf{D2616}, \textbf{D96} \\
\cmidrule(lr){2-3}
& MIPO  & \textbf{D101}, D128, \textbf{D53}, D108, D95, D259, \textbf{D106}, D131, D98, D55 \\
\cmidrule(lr){2-3}
& NECHO & \textbf{D96}, D98, \textbf{D101}, \textbf{D53}, D138, \textbf{D238}, \textbf{D49}, \textbf{D106}, \textbf{D2616}, D663 \\
\bottomrule
\end{tabularx}
}
\caption{Case Study of Next Visit Diagnosis Prediction for a Subject ID of 42129 in MIMIC-III Data. The preceding visit part provides a comprehensive information of a patient on demographics, medical codes, and clinical notes whilst the subsequent visit provides the patient’s real medical codes along with predicted ones by MIPO and NECHO. The accurately predicted codes and their matching ground truths are both in boldface.}
\label{tab-case}
\end{table*}

\subsubsection{Case Study} 
\label{sec:case-study}

To qualitatively evaluate the predictive performance between MIPO \citep{peng2021mipo} and our NECHO, we present a case study (Table \ref{tab-case}) using a patient whose medical history shows a progression from a mitral valve issue to complications after surgery and cardiac rhythm disturbances. In the study, codes are formatted according to the Clinical Classifications Software (CCS) and are sequenced based on their priority, significantly influencing the reimbursement for treatment. We prefix them with "D" to make them appear akin to diagnosis codes.

Notably, our NECHO model accurately predicts 6 out of the top-10 diagnosis, outperforming MIPO, which predicts only 3. Firstly, both successfully identify D53 (Disorders of lipid metabolism), D106 (Cardiac dysrhythmias) and D101 (Coronary atherosclerosis and other heart disease), likely due to these diagnoses being part of the patient's prior medical codes. However, NECHO uniquely predicts D238 (Complications of surgical procedures or medical care), D49 (Diabetes mellitus without complication), D2616 (E Codes: Adverse effects of medical care) and D96 (heart valve disorder) which MIPO fails to identify. 

Additionally, our model predicts D238 and D2616 using multifaceted information of both demographics and notes. D238 should be predicted for two points: 1) the patient was initially hospitalised due to emergency health problem according to demographics, and 2) his notes states visual hallucinations, monitoring for pericardial and pleural effusions. The prediction of D2616 aligns with potential risks associated with mitral valve replacement. On the contrary, MIPO's prediction of D259 (Residual codes; unclassified) and D131 (Respiratory failure; insufficiency; arrest (adult)), which is considered less informative and a simple repetition from previous patient visits. D2 (Septicemia) and D3 (Bacterial infection) are not explicitly mentioned in the patient's history, thus extremely challenging to predict. Hence, this demonstrates the necessity of the effective multimodal fusion strategy for its capability of capturing complementary and unique information in other modalities, verifying the effectiveness of the NECHO. 

Apart from multimodal EHR learning, the content following the "Impression" in the preceding notes is only explicitly found in radiology reports. This indicates the importance of considering all available clinical note types to acquire a thorough understanding of a patient's information. This contrasts with previous findings \citep{hsu2020characterizing, husmann2022importance} suggesting that certain specific note types are representative in EHR learning.

\section{Conclusion} 

Next visit diagnosis prediction is beneficial in AI-driven healthcare applications and has shown remarkable progress. However, the multifaceted and hierarchical properties of EHR data are beyond the consideration for the most of existing studies. To address these limitations, we introduce the novel multimodal EHR modelling framework, NECHO. It effectively aggregates representations from three heterogeneous modalities through meticulously designed multimodal fusion network and the pair of two bimodal contrastive losses in a medical code-centric manner. It also uses parental level information of ICD-9 codes to regularise each modality-specialised encoder to learn more general information. Experimental results including the ablation studies and case study on MIMIC-III data highlight the NECHO's efficacy and superiority. 

\section{Limitations}
Whilst our proposed framework demonstrates promising advancements in multimodal EHR modelling for next visit diagnosis prediction, it is not without its limitations. 

From a data perspective, firstly, the model's predictions are heavily biased to the training data. This means there's a potential risk that the model might underperform when encountering patterns that is nonexistent in the dataset or originating from the different healthcare settings. Secondly, it operates under the assumption that all data modalities are readily and consistently available for every patient. However, this assumption is impractical in that the availability of data can be compromised due to device malfunctions or human errors. Additionally, from a model perspective, the framework’s applicability is confined and has not been extended to a variety of clinical event prediction tasks, such as mortality, re-admissions, and length of stay, where different modalities might take main status.  

We hope to mitigate aforementioned challenges in the near future, enhancing NECHO's adaptability in real-world clinical scenarios.

\section*{Acknowledgement}
We highly appreciate anonymous EACL reviewers and area chairs for their valuable comments that helped us to enhance quality and completeness of this manuscript. 

\bibliography{references}
\bibliographystyle{acl_natbib}

\newpage

\appendix

\section{Modality-Specific Feature Extraction Modules}
\label{sec:method-modalityspecific}

\subsection{Feature Extraction Module for Medical Codes}
\label{sec:method-code}

Medical codes, particularly those from ICD-9 codes, play a vital role in that they directly indicate a patient's status. They are highly specific, unambiguous and succinct, thus they have acted as a primary modality for next admission diagnosis prediction and shown better performance than models leveraging other modalities. Hence, here in this task, we consider them as a main modality.

We employ a single embedding layer $ \text{{E}}_\text{C} $ to process a set of diagnosis codes at $ t $-th patient record, $ c_t $.  The features are passed to a single linear layer followed by a ReLU activation function. It is formulated as:
\begin{equation}
    \bar{c}_t = \text{{E}}_\text{C}(c_t),
\end{equation}
\begin{equation}
    \bar{C}_t = \text{ReLU}(\text{Linear}(\bar{c}_t))
\end{equation}

where $ \bar{C}_t $ represents a feature vector from medical code information of each patient $ \mathcal{P} $ at $t$-th visit.

\subsection{Feature Extraction Module for Demographics}
\label{sec:method-demo}

Each patient has unique demographics, such as gender, age, admission and discharge location, to just name a few. Those provide the supplementary but highly personalised information, allowing an improvement in predictive performance. 

We capture the non-stationary nature of the aforementioned attributes across clinical records at the individual level. For example, variables such as age and insurance type may change over time. Thus, we employ a single embedding layer $ \text{{E}}_\text{H}^{n} $ to $n$-th attribute at $t$-th patient record, $ h_t^n $. The features from each embedding layer are then concatenated and fed into a single linear layer paired with a ReLU activation function. It can be represented as:
\begin{equation}
    \bar{h}_t = \text{concat}(\text{{E}}_\text{H}^{1}(h^1_t) ; \text{{E}}_\text{H}^{2}(h^2_t) ; \cdots ; \text{{E}}_\text{H}^{n}(h^n_t)),
\end{equation}
\begin{equation}
    \bar{H}_t = \text{ReLU}(\text{Linear}(\bar{h}_t))
\end{equation}

where $ \bar{H}_t $ represents a feature vector from demographics of each patient $ \mathcal{P} $ at $t$-th visit.

\subsection{Feature Extraction Module for Clinical Notes}
\label{sec:method-note}

Clinical notes inherently possess a free, unstructured format but carry a comprehensive insight into a patient's condition from the perspective of healthcare provider. They offer potential diagnoses and planned procedures, providing complementary and supplementary information not explicitly specified in medical codes. 

We leverage a combination of pre-trained BioWord2Vec \citep{zhang2019biowordvec} (frozen during both training and inference) and 1D CNN \citep{kim2014convolutional}, which is capable of processing more tokens with computational efficiency. Although many preceding studies utilise PLMs like Clinical BERT \citep{alsentzer2019publicly}, they are still limited by a 512-token maximum, preventing themselves from processing an entire note in a single visit. Thus, we do not utilise them here.

First, we combine all notes $ W_t^1, W_t^2, \ldots, W_t^K $ in a single patient visit $ V_t $ to generate a single note $ W_t $. Then, using the pre-trained BioWord2Vec \citep{zhang2019biowordvec} $ \text{E}_\text{W} $, each discrete word $ w_t^n $ in the note $ W_t $ is mapped to a low-dimensional embedding space, generating $ e_t^n $. With the maximum number of words $ \vert \mathbb{W} \vert $, the word embeddings $ e_t = (e_t^1, e_t^2, \ldots, e_t^{\vert \mathbb{W} \vert}) $ from the combined note $ W_t $ are then fed into the 1D CNN (\text{Conv1D}) with multiple filters with a subsequent max-pooling layer ($ \text{Max} $) to generate the most salient features $ \bar{w}_t $ using a filter (equivalent to window size) $ f $. The outputs from each filter are concatenated and passed to a linear layer with ReLU activation function. It yields the note representation $ \bar{W}_t $ at $t$-th visit of each patient $ \mathcal{P} $. The aforementioned processes are mathematically described as follows:
\begin{equation}
    W_t = \text{concat}(W_t^1 ; W_t^2 ; \cdots ; W_t^K),
\end{equation}
\begin{equation}
    e_t^n = \text{E}_\text{W}(w_t^n),
\end{equation}
\begin{equation}
\begin{aligned}
    \bar{e}_t^{f} = \text{ReLU}((\text{Conv1D}^{f}(e_t)) \\
    \text{where} \: f \in [2, 3, 4],
\end{aligned}
\end{equation}
\begin{equation}
    \bar{w}_t^{f} = \text{Max}(\bar{e}_t^{f}),
\end{equation}
\begin{equation}
    \bar{w}_t = \text{concat}(\bar{w}_t^{2} ; \bar{w}_t^{3} ; \bar{w}_t^{4}),
\end{equation}
\begin{equation}
    \bar{W}_t = \text{ReLU}(\text{Linear}(\bar{w}_t)).
\end{equation}

\newpage

\section{Data Pre-processing} 
\label{sec:appendix-data}

\textbf{Patient Selection Criteria} We follow the previous work of GRAM \citep{choi2017gram}. First, we select patients with minimum two visits. Also, we truncate visits beyond the 21st visit.

\textbf{Demographics Processing} Attributes such as age, gender, admission type, admission and discharge locations, and insurance type are considered. Patients with ages 0 or above 120 are excluded. The admission types encompass categories such as emergency, elective, and urgent whilst the insurance types include medicare, private, medicaid, government and self pay. The dataset also offers a diverse range of features for both admission and discharge locations.

\textbf{Clinical Note Processing} Even though some prior works \citep{hsu2020characterizing, husmann2022importance} emphasise the significance of specific note types for EHR representation learning, we consider all available note types (e.g. radiology, discharge summary, and nursing) for universality.

We first pre-process the notes, following the previous work \citep{khadanga2019using}. It involves a removal of non-alphabetical characters, stopwords and conversion of uppercase to lowercase letters. Then, we add two special tokens to BioWord2Vec \citep{zhang2019biowordvec}, <PAD> and <UNK>, the same as those used in BERT \citep{devlin2018bert}. They are initialised using matrices filled with zeros and uniform distribution, respectively. Any visit records lacking note information are excluded. Next, each note is tokenised with maximum 10k words using BioWord2Vec. This approach effectively captures the entirety of note information for approximately 85\% of all the visits. 

\textbf{Medical Ontology \& Label Construction} Following the GRAM \citep{choi2017gram}, a medical ontology is constructed based on ICD-9 codes using the Clinical Classifications Software (CCS) from the Healthcare Cost and Utilization Project\footnote{https://hcup-us.ahrq.gov/toolssoftware/ccs/ccs.jsp}. The labels are derived from nodes present in the primary\footnote{https://hcup-us.ahrq.gov/toolssoftware/ccs/AppendixCMultiDX.txt} and secondary\footnote{https://hcup-us.ahrq.gov/toolssoftware/ccs/AppendixASingleDX.txt} hierarchy of the ICD-9 codes. This renders the next visit diagnosis prediction task as a hierarchical multi-label multi-class classification. 

\textbf{Summary} A comprehensive statistical summary of the pre-processed dataset is provided in Table \ref{tab-stats}. 

\newpage
\begin{table}[h]
\resizebox{\columnwidth}{!}{%
\begin{tabular}{lc}
    \hline
    \hline
    Dataset & MIMIC-III\\
    \hline
    \hline
    \# of patients & 6,812 \\
    \# of visits &  18,256 \\
    Avg. \# of visits per patient & 2.68 \\
    \hline
    \# of Training Data & 5449 \\
    \# of Validation Data & 681 \\
    \# of Test Data & 682 \\
   \hline
    \# of unique ICD9 codes &  4,138 \\
    Avg. \# of ICD9 codes per visit & 13.27 \\
    Max \# of ICD9 codes per visit & 39 \\
    \hline
    \# of category codes & 265 \\
    Avg. \# of category codes per visit & 11.40 \\
    Max \# of category codes per visit & 34 \\
    \hline
    \# of disease typing code & 17 \\
    Avg. \# of disease typing codes per visit & 6.68 \\
    Max \# of disease typing codes per visit & 15 \\
    \hline
    \# of Age & 73 \\
    \# of Gender & 2 \\
    \# of Admission Type & 3 \\
    \# of Admission Location & 8\\
    \# of Discharge Location & 16 \\
    \# of Insurance Type & 5\\
    \hline
    Avg. \# of words per visit & 6743 \\
    Max \# of words per visit & 239,102 \\
    \hline
\end{tabular}%
}
\caption{Statistics of the Pre-processed MIMIC-III Data.}
\label{tab-stats}
\end{table}

\newpage

\section{Experiments on the Coefficient for Hierarchical Regularisation} 
\label{sec:appendix-hierarchy}

We assume that modality-specific encoders necessitate soft regularisation for two reasons: firstly, their representations are relatively incomplete in comparison to the full framework (NECHO); secondly, since the general information embodies a broader scope, it should not impose excessive constraints on these encoders during training. 

The empirical results on Table \ref{tab-hierreg}, delineated on a logarithmic scale for $ \lambda_{\text{hrchy}} $ values ranging from 0.01, 0.1, to 1, substantiate our hypothesis. Notably, setting it as 0.1 enhances the overall model performance the most, thereby verifying its optimal effectiveness.

\begin{table}[h!]
\centering
\resizebox{0.35\textwidth}{!}{
\begin{tabular}{ccccc}
\toprule
\midrule
\multirow{2}{*}{\textbf{Coefficients}} & \multirow{2}{*}{\textbf{Values}} & \multicolumn{2}{c}{\textbf{Acc@$\textbf{k}$}}   \\
\cmidrule(lr){3-4}
&        & \textbf{10}    & \textbf{30} \\ 
\midrule 
\midrule
\multirow{3}{*}{$ \lambda_{\text{hrchy}} $} & 0.01    & 42.24    & 70.09      \\
   &  0.1   & \textbf{43.55}     & \textbf{71.45}       \\
   &  1  & 43.02    & 70.82      \\    
\bottomrule
\end{tabular}%
}
\caption{Experimental Results on MIMIC-III Data of the Coefficient for Hierarchical Regularisation, $ \lambda_{\text{hrchy}} $.}
\label{tab-hierreg}
\end{table}

\newpage

\section{Baselines} 
\label{sec:appendix-baselines}

\subsection{Unimodal EHR Modelling Baselines}

\begin{itemize}
    \item GRAM \citep{choi2017gram} considers medical ontology with an attention mechanism.
    \item KAME \citep{ma2018kame} employs an attention mechanism at the knowledge level, specifically tailored for medical ontology.
    \item MMORE \citep{song2019medical} attentively learns both the multiple ontological representation and the co-occurrence statistics.    
    \item MIPO \citep{peng2021mipo} utilises an auxiliary task of disease typing task. In other words, it learns parental level ICD-9 codes additionally.    
    \item Medical Code Encoder (Ours) employs a simple combination of embedding layers and a couple of linear layers, which are followed by ReLU and Sigmoid activation function. It is utilised in our pipeline. Refer to Appendix \ref{sec:method-code} for details.
    \item Demographics Encoder (Ours) utilises a simple combination of attribute-specific embedding layers and two linear layers, whose subsequent layers are ReLU and Sigmoid activation function, respectively. It is employed in our pipeline. Refer to Appendix \ref{sec:method-demo} for details.
    \item BioWord2Vec \citep{zhang2019biowordvec} model is combined with 1D CNN \citep{kim2014convolutional}. For brevity, we simplify it as BioWord2Vec. It uses pre-trained embedding with 16,545,454 words (with an arbitrary addition of two special tokens), which are subsequently processed by 1D CNN. In our framework, this serves as the notes feature extraction module. Refer to Appendix \ref{sec:method-note} for details. 
    \item Bio-Clinical BERT \citep{alsentzer2019publicly} is a derivative of the original BERT \citep{devlin2018bert} on bio-medical domain. It is trained on MIMIC-III dataset \citep{johnson2016mimic} and has a maximum input sequence length of 512. 
\end{itemize}

\subsection{Multimodal EHR Modelling Baselines} 

Both MNN and MAIN process 10k words from a clinical note within a single visit. The parameters (e.g. hidden dimension, the number of heads) are set in accordance with the specifications detailed in their original paper.

\begin{itemize}
    \item MNN \citep{qiao2019mnn} is trained using both medical codes and clinical notes. It employs a single embedding layer for the former and a combination of BioWord2Vec 1D CNN for the latter. The fusion of representations from these two modalities is achieved through deep feature mixture \citep{lian2018xdeepfm} and bi-directional RNN with attention. 
    \item MAIN \citep{an2021main} is a trimodal model, integrating medical codes, clinical notes, and demographics, which is akin to our approach. First, medical codes and clinical notes are fused using a combination of low-rank fusion \citep{liu2018efficient} and cross-modal attention. Next, demographics is merged using low-rank fusion subsequently. 
\end{itemize}

\subsection{Multimodal Fusion Strategies Baselines}

We employ the same feature extraction module as used in our approach for the subsequent baselines, and fuse different modalities using their proposed mechanisms. For fairness, we set the parameters as the same as ours.

\begin{itemize}
    \item Concat, an abbreviation for concatenation, is a straightforward method that merges distinct modalities without any computations, ensuring a raw and unaltered integration.
    \item TFN (Tensor fusion Network) \citep{zadeh2017tensor} executes an outer product on the representations of different modalities.     
    \item MulT (Multimodal Transformer) \citep{tsai2019multimodal} utilises both cross-modal and self-attention transformers to integrate distinct modalities.
    \item MAG (Multimodal Adaptation Gate) \citep{rahman2020integrating} refines the representation of one modality by adjusting it with a displacement vector, which is derived from the other modalities. 
    \item ULGM (Unimodal Label Generation Module) \citep{yu2021learning} uses modality-specific encoders to predict the ground truths as well.
    \item TeFNA (Text Enhanced Transformer Fusion Network) \citep{huang2023tefna} learns text-centric pairwise cross-modal representations.
\end{itemize}

\newpage

\section{A Comparative Study on Different MAGs} 
\label{sec:appendix-MAG}

We present a comparative analysis of various MAGs, including our newly developed code-centric MAG and others \cite{rahman2020integrating, yang2021leverage}. \cite{rahman2020integrating} introduce MAG initially while MAG from \cite{yang2021leverage} combines representations from different modalities at the sample level dynamically with an attention gate. They are replaced with our MAG in the framework for a comparison.

From the Table \ref{tab-MAGs}, it demonstrates the superiority of our method over preceding approaches. It can be attributed to the meticulous consideration of the modality imbalance, one of factors not adequately addressed by previous methodologies. This validates that considering the dominance of main modality is essential in multimodal modelling.

\begin{table}[h]
\centering
\resizebox{0.43\textwidth}{!}{
\begin{tabular}{ccccc}
\toprule
\midrule
\multirow{2}{*}{\textbf{Criteria}} & \multirow{2}{*}{\textbf{Methodologies}} & \multicolumn{2}{c}{\textbf{Acc@$\textbf{k}$}}   \\
\cmidrule(lr){3-4}
&        & \textbf{10}    & \textbf{30} \\ 
\midrule 
\midrule
\multirow{3}{*}{MAG} & \cite{rahman2020integrating}    & 42.36    & 69.16      \\
   &  \cite{yang2021leverage}   & 42.24     & 70.22       \\
   &  NECHO (Ours) & \textbf{43.55}    & \textbf{71.45}      \\    
\bottomrule
\end{tabular}%
}
\caption{Experimental Results on MIMIC-III Data on Different MAGs.}
\label{tab-MAGs}
\end{table}

\end{document}